\documentclass[journal, hidelinks, 10pt]{IEEEtran}

\IEEEoverridecommandlockouts                              

\usepackage{bbm}
\usepackage{graphicx}
\usepackage{amsmath}
\usepackage{caption}
\usepackage{subcaption}
\usepackage{amsfonts}
\usepackage{amssymb}
\usepackage{algorithm,algorithmic}
\usepackage{cite}
\usepackage{array}
\usepackage{orcidlink}
\usepackage{soul,color}
\usepackage{multirow}
\usepackage{hyperref}[hidelinks]
\usepackage{soul}
\usepackage{dsfont}
\usepackage{fancyhdr}

\newcolumntype{M}[1]{>{\centering\arraybackslash}m{#1}}

\title{\LARGE \bf
Intelligent Mode-switching Framework for Teleoperation}

\author{Burak Kizilkaya$^{1}$, Changyang She$^{2}$, Guodong Zhao$^{1}$, Muhammad Ali Imran$^{1}$
\thanks{$^{1}$Burak Kizilkaya, Guodong Zhao, and Muhammad Ali Imran are with James Watt School of Engineering, University of Glasgow, Glasgow G12 8QQ, U.K.
        {\tt\small Email: burak.kizilkaya@glasgow.ac.uk, guodong.zhao@glasgow.ac.uk, \\ muhammad.imran@glasgow.ac.uk }}%
\thanks{$^{2}$Changynag She is with School of Electrical and Information Engineering, The University of Sydney, Sydney, NSW 2006, Australia.
        {\tt\small Email: shechangyang@gmail.com}}%
}

\begin{document}

\maketitle

\begin{abstract}
Teleoperation can be very difficult due to limited perception, high communication latency, and limited degrees of freedom (DoFs) at the operator side. Autonomous teleoperation is proposed to overcome this difficulty by predicting user intentions and performing some parts of the task autonomously to decrease the demand on the operator and increase the task completion rate. However, decision-making for mode-switching is generally assumed to be done by the operator, which brings an extra DoF to be controlled by the operator and introduces extra mental demand. On the other hand, the communication perspective is not investigated in the current literature, although communication imperfections and resource limitations are the main bottlenecks for teleoperation. In this study, we propose an intelligent mode-switching framework by jointly considering mode-switching and communication systems. User intention recognition is done at the operator side. Based on user intention recognition, a deep reinforcement learning (DRL) agent is trained and deployed at the operator side to seamlessly switch between autonomous and teleoperation modes. A real-world data set is collected from our teleoperation testbed to train both user intention recognition and DRL algorithms. Our results show that the proposed framework can achieve up to 50\% communication load reduction with improved task completion probability.
\end{abstract}

\section{Introduction}\label{intro}
Achieving real-time teleoperation\footnote{Teleoperation covers any remote operation done by a human operator by controlling a remote robot (which can be a manipulator robot, mobile robot, UGV, UAV, etc.) over a communication network.}, in which a human operator manually controls a remote teleoperator, is challenging and requires an excessive amount of efforts to provide satisfactory user experience due to the limited perception, high communication latency, and limited degrees of freedom (DoFs) at the operator side \cite{zein2020enhanced}. In addition, the existing teleoperation systems can hardly meet the mental and physical demands as the operator needs to concentrate on every single detail in the control process instead of focusing on the task at hand \cite{zein2021deep}. To address these issues, autonomous teleoperation systems were developed in the existing literature\cite{tanwani2017generative, akay2022end, herlant2016assistive, gao2014contextual, zein2020enhanced, zein2021deep, gonzalez2021deserts, agarwal2021dexterous, hussein2021personalized, hussein2022incremental, jain2019probabilistic}. The basic idea of autonomous teleoperation is to predict user intention and execute some parts of the task autonomously to decrease the demand on the operator and increase the task completion rate. Existing studies can be categorized based on user intention recognition techniques, task performance metrics, and decision-making algorithms. User intention recognition is performed either by model based methods\cite{tanwani2017generative, gao2014contextual, jain2019probabilistic}, data-driven methods \cite{zein2020enhanced, zein2021deep, gonzalez2021deserts, agarwal2021dexterous, hussein2021personalized, hussein2022incremental} or combinations of both types of methods \cite{akay2022end}. It is worth noting that user intention recognition accuracy varies between 20\% and 95\%, which was not considered in some references. Performance metrics highly depend on specific tasks. To better measure the performance, both objective metrics (e.g., task success rate and task completion time) and subjective metrics (e.g., operators' mental and physical demands) are considered.
In some existing studies\cite{zein2020enhanced, zein2021deep}, decision-making for mode-switching is assumed to be done by the operator. This approach brings an extra DoF to be controlled by the operator and generally introduces extra mental demand \cite{akay2022end}. Hence, developing a mode-switching policy that works autonomously and seamlessly is in urgent need. Furthermore, communication delay and resource limitations are the main bottlenecks for long-distance teleoperation. Nevertheless, how a communication system can be designed to accommodate autonomous teleoperation efficiently remains an open question. Therefore, we need a new design methodology to jointly consider autonomous teleoperation systems and communication systems to achieve seamless operation with low physical and mental operator demand.

With the aforementioned considerations, we propose an intelligent mode-switching framework for long-distance teleoperation. We design a general framework by jointly considering the communication system and the decision-making system. User intention recognition is done at the operator side with a CNN-based classification model. Based on user intentions, a deep reinforcement learning (DRL) agent is trained and deployed on the operator side to switch between the autonomous and teleoperation modes. A real-world data set is collected from our teleoperation testbed to train both user intention recognition and DRL algorithms. Our results show that the proposed framework can achieve up to 50\% communication load reduction and improve task completion probability.

\section{System Model}

\begin{figure}
    \centering
    \includegraphics[width=0.47\textwidth]{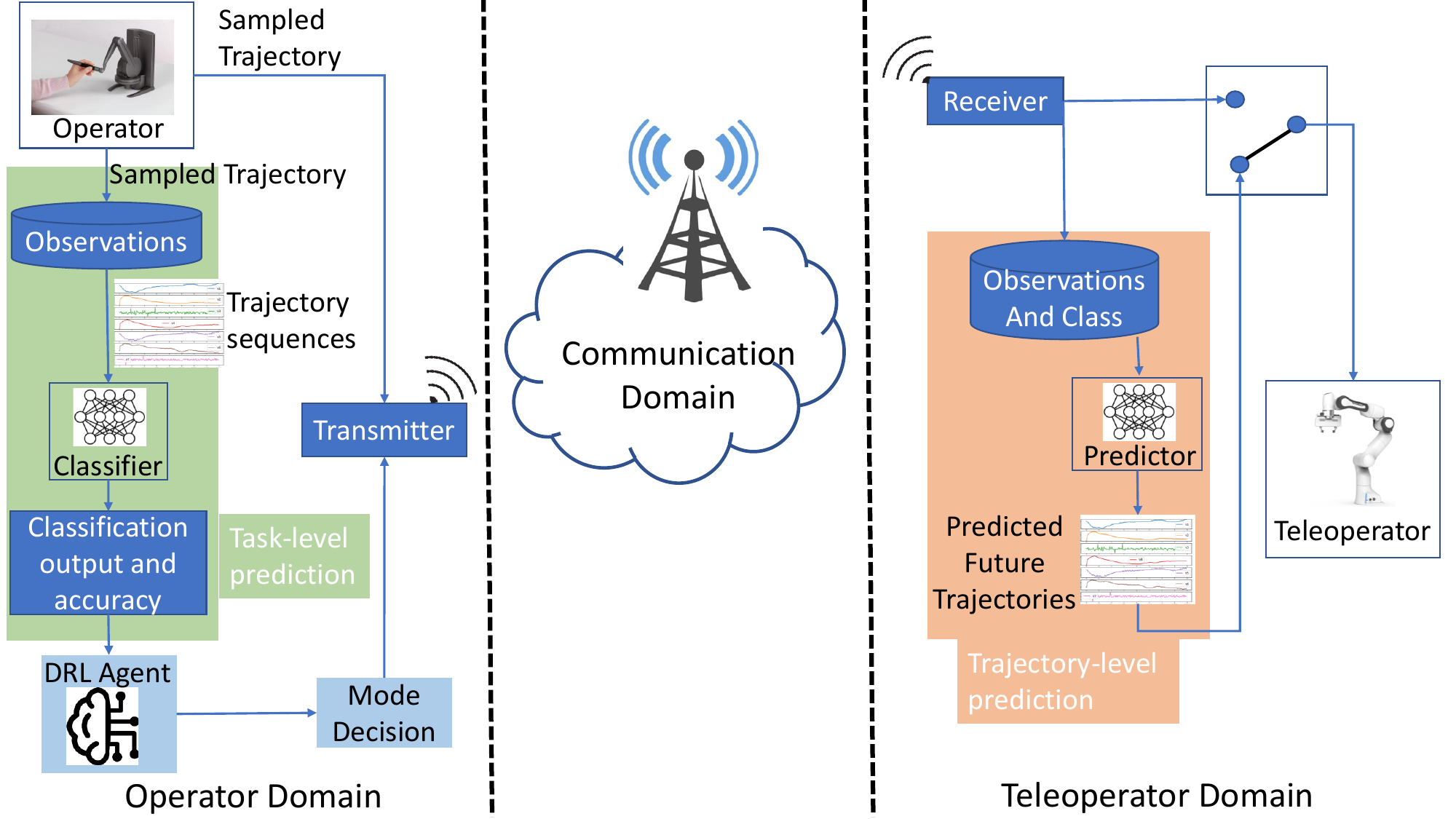}
    \caption{Intelligent Mode Switching Framework}
    \label{fig:local_system}
\end{figure}

The proposed framework consists of three main domains, namely operator domain, communication domain, and teleoperator domain, as shown in Fig. \ref{fig:local_system}, and time is discretized into slots. A human operator uses a controller with a haptic interface to control a remote teleoperator. At time slot $t$, $m$-th operator's control commands, $K_m(t)$, are sampled and transmitted. Sampled control commands are also saved in the observation memory to be used by a local task-level prediction algorithm, which attempts to recognize the user's intention in every time slot using the history of observations. Prediction output, $L_m(t) = \{l_{m}^1(t), ..., l_{m}^{N}(t)\}^{\rm{T}}$, is a probability vector obtained from task-level prediction algorithm, where $l_{m}^n(t) \in [0,1]$ is the probability that the user intends to execute task $n$ in the $t$-th time slot. $L_m(t)$ is used by a DRL agent to make a mode-switching decision, and we will introduce the intention recognition algorithm and the DRL algorithms in the next section. In teleoperation mode, the system works as a conventional teleoperation system in which the controller and the teleoperator exchange packets over the communication domain. In autonomous mode, there is no real-time packet exchange between the operator and the teleoperator. Task-level predictor works as an error detection mechanism in autonomous mode, which enables recovery from wrong user intention recognition. In case of prediction error, the system switches back to the teleoperation mode, i.e., gives control to the operator, and continues to monitor prediction output to recognize user intention for possible switching to autonomous mode again. Teleoperator is also equipped with a trajectory-level predictor that is used in autonomous mode. In other words, the teleoperator predicts future trajectories and finishes the task autonomously. It is worth noting that the computation time for user intention recognition and error detection is less than 1 ms considering execution time of a CNN-based classification model \cite{shustanov2017cnn}. Therefore, we assume that it is negligible compared to long-distance communication delay.

\subsection{Communication Load}
We define the communication load as the average data rate required for a task, $D_m$, which can be given as
\begin{align}
    D_m = \frac{d_m}{Z_m} b_m\; (bits/slot),
\end{align}
\noindent where $d_m$ is the number of time slots the controller is in the teleoperation mode, $Z_m$ is the number of time slots required to finish the task and $b_m$ is the number of bits transmitted in each time slot. We consider 5G New Radio (NR) as an example communication system. For teleoperation applications, the packet size is small and the transmission duration of each packet is much smaller than the channel coherence time. In other words, the channel fading coefficient remains constant over the transmission duration \cite{durisi2016toward}. In such a scenario, the maximal achievable rate can be accurately approximated as \cite{yang2014quasi}
\begin{equation}
    B_m  \approx C_m - \sqrt{\frac{V_m}{\tau_m W_m }} Q^{-1}(\epsilon^\text{d}_m)\; \text{(bits/s/Hz),}
\end{equation}
\noindent where $C_m = \log(1+\gamma_m)$ is the Shannon capacity, $\gamma_m = \frac{\alpha_m g_m P_m}{N_0W_m}$ is the received signal to noise ratio (SNR) at the base station, $\alpha_m$ is the large-scale channel gain, $g_m$ denotes the small-scale channel gain, $P_m$ denotes the transmit power, $N_0$ is the single sided noise spectral density, $V_m = \log(e)^2 \left[1 - \frac{1}{(1+\gamma_m)^2}\right]$ is the channel dispersion, $\tau_m W_m $ is the blocklength, $\tau_m$ is the transmission duration, $W_m$ is the bandwidth, $Q^{-1}(.)$ is the inverse of the Gaussian Q-function, $Q(x) \triangleq \int_{x}^{\infty} \frac{1}{2\pi}e^{-t^{2}/2} dt$, and $\epsilon^\text{d}_m$ is the decoding error probability. Given that $b_m = \tau_m W_m B_m$, communication load, $D_m$ becomes,
\begin{align}
    D_m \approx \frac{d_m \tau_m W_m}{Z_m} \left[C_m - \sqrt{\frac{V_m}{\tau_m W_m }} Q^{-1}(\epsilon^\text{d}_m)\right]\; (bits/slot).
\end{align}
\subsection{Task Completion Probability}
The task completion probability can be analyzed in teleoperation mode and autonomous mode. We denote the probability that the system stays in the teleoperation mode and the autonomous mode by $P^{\rm t}$ and $P^{\rm a}$, respectively, where $P^{\rm t} + P^{\rm a} = 1$.
\begin{itemize}
    \item \textbf{Case 1: } \textit{Teleoperation Mode}\\
    In the teleoperation mode, the task completion probability highly depends on the communication reliability and the operator's experience since real-time packet exchange takes place. The reliability of a communication system can be measured by the decoding error probability, $\epsilon^{\rm d}_m$, and the queuing delay violation probability, $\epsilon^{\rm q}_m$. To take operator experience into account, we denote the operator experience coefficient by, $\rho_m\in[0,1]$. In this case, task completion probability becomes the multiplication of the communication reliability and user experience coefficient. In other words, any communication imperfections such as decoding error or queuing delay violation, and operator errors can cause task failure. Hence, the task completion probability in the teleoperation mode can be expressed with the following relationship between the decoding error probability, $\epsilon^{\rm d}_m$, the queuing delay violation probability, $\epsilon^{\rm q}_m$ and the operator experience coefficient $\rho_m$.
    \begin{align}
        P^{\rm \mu}_m & =  f^{\rm \mu}_m(\epsilon^{\rm q}_m, \epsilon^{\rm d}_m, \rho_m) \\
        &  = (1-\epsilon^{\rm q}_m) (1 - \epsilon^{\rm d}_m) \rho_m \nonumber
        \label{eq_succ_tele}
    \end{align}

    \item \textbf{Case 2: } \textit{Autonomous Mode}\\
    In the autonomous mode, task completion probability depends on the intention recognition algorithm, the error detection algorithm, and the trajectory prediction algorithm. To take these into account, we define the task-level prediction error probability, $\epsilon^{ \rm c}_m$, the error detection system failure probability, ${\epsilon^{ \rm f}_m}$, and the trajectory-level prediction error probability, $\epsilon^{\rm t}_m$. In this case, task completion probability becomes the multiplication of the task-level prediction reliability, error detection system reliability and trajectory-level prediction reliability. In other words, prediction error on trajectory-level prediction can cause task failure. Similarly, prediction error in task-level can cause task failure if it cannot be captured by error detection system. Hence, task completion probability in the autonomous mode can be expressed with the following relationship between the task-level prediction error probability, $\epsilon^{ \rm c}_m$, the error detection system failure probability, ${\epsilon^{ \rm f}_m}$, and the trajectory-level prediction error probability, $\epsilon^{\rm t}_m$.
    \begin{align}
        P^{\rm \sigma}_m & =  f^{\rm \sigma}_m(\epsilon^{\rm c}_m, \epsilon^{\rm f}_m, \epsilon^{\rm t}_m)\\
        & = ((1 - \epsilon^{\rm c}_m) + (\epsilon^{\rm c}_m(1 - \epsilon^{\rm f}_m)))(1-\epsilon^{\rm t}_m) \nonumber
    \end{align}
Then, the overall task completion probability can be given as,
\begin{align}
    P^{\rm o}_m & = P^{\rm t} P^{\rm \mu}_m + P^{\rm a} P^{\rm \sigma}_m, \nonumber\\
    & = P^{\rm t} f^{\rm \mu}_m(\epsilon^{\rm q}_m, \epsilon^{\rm d}_m, \rho_m) + P^{\rm a} f^{\rm \sigma}_m(\epsilon^{\rm c}_m, \epsilon^{\rm f}_m, \epsilon^{\rm t}_m). 
    \label{eq_succ_prop}
\end{align}

\end{itemize}

\subsection{Problem Formulation}
To efficiently use available wireless resources, we minimize communication load, $D_m$, by jointly optimizing communication and mode-switching systems subject to joint task completion probability, $P^{o}_m$, which can be formulated as
\begin{align}
    \min_{b_m, d_m}
    & \; {D_m \approx \frac{d_m \tau_m W_m}{Z_m} \left[C_m - \sqrt{\frac{V_m}{\tau_m W_m }} Q^{-1}(\epsilon^\text{d}_m)\right]\;} \label{generalObjective}\\
    \text{s.t.}\; & \;
    P^{o}_m > \psi_m. \nonumber
\end{align}
\noindent where $\psi_m$ is a task-dependent task completion requirement.

\section{Intelligent Mode-switching Framework}
\subsection{Task-level Prediction: User Intention Recognition}
\label{sec-task-level-pred}
User intention recognition problem can be modeled as time series classification where Convolutional Neural Networks (CNNs) \cite{zhao2017convolutional}, Long Short Term Memory (LSTM) \cite{karim2017lstm}, Recurrent Neural Networks (RNNs), and distance based classification methods \cite{abanda2019review} such as k-Nearest Neighbours (kNNs) and Dynamic Time Warping (DTW) are used in the existing literature \cite{ismail2019deep}. In this study, we design CNN based time series classification model. In input layer, multivariate time series data, $\textbf{k}_{t:t^{'}}$, are received at time slot $t^{'}$ with observation length $t^{'}-t$ time slots. In convolution layers, the element-wise convolution operation is applied to compute the feature representations of inputs (i.e., \textit{feature maps}). For input $\textbf{k}_{t:t^{'}}$ and kernel $\varrho_t$, the resulting feature at location $(i,j)$ can be computed as,
\begin{align}
&\textbf{X}_{t:t^{'}}^{i,j} = \textbf{W}_{\varrho_{t:t^{'}}} * \textbf{k}_{t:t^{'}}^{i,j} + \textbf{b}_{\varrho_{t:t^{'}}},\\
&\textbf{Y}_{t:t^{'}}^{i,j} = \Phi(\textbf{X}_{t:t^{'}}^{i,j}),
\end{align}
\noindent where $\textbf{W}_{\varrho_{t:t^{'}}}$ and $\textbf{b}_{\varrho_{t:t^{'}}}$ are the weights and bias of the filter $\varrho_{t:t^{'}}$, $\textbf{k}_{t:t^{'}}^{i,j}$ is the subsection of the input centered at $(i,j)$, $\Phi(\cdot)$ is the non-linear activation function, and `$*$' is the convolution operator. Then, activation layers are employed after convolution layers using Leaky Rectified Linear Unit, \textit{LeakyReLu}, activation function. After feature map computations with several convolution and activation layers, the pooling layer is employed to decrease the dimension of feature maps in which global average pooling is used. Then, two dense, i.e., fully connected, layers are employed to train the classification model with \textit{softmax} activation function. For a given observation window, $\textbf{k}_{t:t^{'}}$, the classifier is trained to predict the class of the task. The categorical cross-entropy function is used as a loss function with one-hot encoded labels. Each convolution layer has 128 CNN cells and the batch size is 64. Early stopping criteria is adopted to avoid over-fitting, i.e., the best model is saved if there is no improvement for 10 consecutive epochs in the training process. 
\subsection{Trajectory-level Prediction}
\begin{table}
 \caption{Hyper-Parameters of Trajectory-level Prediction Models}
    \centering
    \begin{tabular}{|M{3cm}|M{2cm}|M{2.3cm}|}
    \hline
        &\textbf{LSTM} & \textbf{CNN}\\
        \hline
        \textbf{Number of layers} &  1  & 1  \\
        \hline
        \textbf{Number of cells}& 128 LSTM cells & 128 CNN cells\\
        \hline
        \textbf{Batch size}& 128 &  128\\
        \hline
        \textbf{Optimizer}&Adam  & Adam \\
        \hline
        \textbf{Loss function}& MSE  & MSE \\
        \hline
        \textbf{Accuracy metric}& RRMSE & RRMSE \\
        \hline
        \textbf{Max training epochs}&   1000 & 1000\\
        \hline
         \textbf{Activation function}& tanh  & ReLu \\
        \hline
    \end{tabular}
    \label{tab:pred_models}
\end{table}
The teleoperator predicts the future trajectory from the observed trajectory to be able to finish the task autonomously, in the autonomous mode. Let's assume the teleoperator has observation of trajectories, $\textbf{k}_{t:t^{'}}$, at $t^{'}$-th time slot with observation length $t^{'}-t$. Given the observation window and the task label, the trajectory-level predictor predicts future trajectory, $\textbf{k}_{t^{'}:t^{''}}$, with prediction horizon $t^{''}-t^{'}$. We use two types of predictors, namely LSTM and CNN for trajectory-level prediction. Observation and prediction window lengths depend on mode-switching timings. For example, if the mode is switched from teleoperation to autonomous in the middle of the task, then the observation and the prediction windows are in equal length. Hyper-parameters of trajectory-level prediction algorithms are given in Table \ref{tab:pred_models}.

\subsection{DRL Framework}
In this section, we formulate the problem as discrete-time Markov Decision Process (MDP).
\subsubsection{States}
Let's denote the state of the $M$ devices in the $t$-th slot as $\textbf{S}(t) = (S_1(t), S_2(t),..., S_M(t))$. The state of the $m$-th device consists of the predicted user intention, $L_m(t)$, the observation length, $T_m(t)$, and the current mode, $M_m(t)$, i.e., 
\begin{align}
    S_m(t) = \{L_m(t), T_m(t), M_m(t)\}
\end{align}
\subsubsection{Actions}
The action to be taken by the controller is to either switch the mode or keep the system in the same mode. We denote the action by $A_m(t)$, which is a discrete binary value: (0) staying in the current mode (1) switching to a different mode.
\subsubsection{Reward}
DRL agent errors can have significant consequences, impacting both the communication load and the successful completion of tasks. Firstly, when a DRL agent remains in teleoperation mode instead of switching to autonomous mode, it can result in unnecessary communication load. This inefficiency can hinder the overall system performance. Secondly, if the DRL agent mistakenly switches to autonomous mode when it should remain in teleoperation mode, it may lead to task failure. To address these challenges effectively, we have defined a reward function that considers these potential pitfalls. By providing a small reward in autonomous mode, we encourage the DRL agent to minimize communication load. Additionally, we incorporate an ultimate reward, contingent upon task completion outcomes, to encourage successful task execution. This reward structure ensures that the DRL agent is motivated to balance both communication efficiency and task accomplishment, thus improving overall system performance and reducing the likelihood of errors. Therefore, the reward depends on the current state and the final outcome of the task, which is defined as follows.
\begin{align}
R_m(t) = 
 \begin{cases} 
      1, & t\leq Z_m\; \text{and}\; M_m(t)=auto \\
      0, & t\leq Z_m\; \text{and}\; M_m(t)=tele \\
      \frac{Z_m-d_m}{Z_m} \times 100, & t=Z_m\; \text{and}\; P^{o}_m > \psi_m  \\
  \end{cases}    
\end{align}
\subsection{DRL Training}
We apply Deep Q-Learning (DQN) \cite{mnih2013playing} to find the optimal mode-switching policy. The details are given in Algorithm \ref{alg1}. The DQN algorithm has a main network, $Q$, and a target network, $\tilde{Q}$. The initial parameters of both neural networks are denoted by $\theta$ and $\tilde{\theta}$, respectively. We apply the $\varepsilon$-greedy policy to achieve a good trade-off between exploration and exploitation, where the agent chooses a random action with probability $\varepsilon$ and chooses the action, $A_t$, with probability $1-\varepsilon$ according to,
\begin{align}
    A_t = \arg\max_{A_t}Q(S_t, A_t | \theta).
\end{align}
After taking the action, $A_t$, agent observes the reward, $R_t$, and the next state, $S_{t+1}$. The observed transition $<S_t, A_t, R_t, S_{t+1,}, \mathds{1}_t>$  is stored in replay buffer, $\Gamma$, with capacity $\kappa$. We denote $\mathds{1}_t$ as an indicator showing whether the task is terminated or not in the $t$-th slot. After collecting enough transitions in the replay buffer, the training starts. Specifically, the agent samples a batch of transitions, $<S_j, A_j, R_j, S_{j+1}, \mathds{1}_j>$, $j \in \mathcal{N}$, and estimates the long-term reward from the target network,
\begin{align}
Y_j = 
 \begin{cases} 
      R_j, & \mathds{1}_j = 1 \\
      R_j + \gamma \max_{A_{j+1}}\tilde{Q}(S_{j+1}, A_{j+1}), & \mathds{1}_j = 0 \\
  \end{cases}    
\end{align}
Then the Stochastic Gradient Descent (SGD) algorithm is applied to minimize the mean square Bellman error, i.e., the difference between the main network and $Y_j$,
\begin{align}
    \mathcal{L} = \frac{1}{N} \sum_{j=1}^{N}(Q(S_j, A_j) - Y_j).
\end{align}
Finally, the target network is updated in every $C$ steps.
\begin{algorithm}
 \caption{DQN training}
 \begin{algorithmic}[1]
 \scriptsize
 \renewcommand{\algorithmicrequire}{\textbf{Input:}}
 \renewcommand{\algorithmicensure}{\textbf{Output:}}
 \REQUIRE Initialize main network, $Q$ and target network, $\tilde{Q}$ with random weights, $\theta$ and $\tilde{\theta}$, respectively. Initialize replay buffer, $\Gamma$. Observe the initial state, $S_t = \{L_t, T_t, M_t\}$, with predicted user intend, $L_t$, observation length, $T_t$, and current mode, $M_t$.\\
 
 \FOR{episode = 1,...}
 \STATE Choose an action, $A_t$, randomly with probability $\varepsilon$, or obtain the action $A_t = \arg\max_{A_t}Q(S_t, A_t | \theta)$ with probability $1-\varepsilon$
  \STATE Apply action $A_t$, observe reward, $R_t$, and next state, $S_{t+1}$
  \STATE Store transition $<S_t, A_t, R_t, S_{t+1}, \mathds{1}_t>$ in replay buffer, $\Gamma$
  \STATE Sample $N$ batch of transitions randomly from replay buffer, $\Gamma$
 \FOR{each transition $<S_j, A_j, R_j, S_{j+1}, \mathds{1}_j>$ in $N$ batches}
  \IF{$done_j$}
  \STATE $Y_j = R_j$
  \ELSE
  \STATE $Y_j = R_j + \gamma \max_{A_{j+1}}\tilde{Q}(S_{j+1}, A_{j+1})$
  \ENDIF
  \ENDFOR 
  \STATE Update $\theta$ by minimising loss $\mathcal{L} = \frac{1}{N} \sum_{j=1}^{N}(Q(S_j, A_j) - Y_j)$
  \STATE Copy $\theta$ into $\tilde{\theta}$ in every $\mathcal{C}$ steps

  \ENDFOR

 \end{algorithmic} 
 \label{alg1}
 \end{algorithm}
\section{Evaluation of the Proposed Framework}
In this section, we evaluate task-level prediction, trajectory-level prediction, and the proposed intelligent mode-switching framework. In the simulations, we use the relationships derived in  (\ref{eq_succ_prop}). In trajectory-level prediction, we utilize the LSTM model. The operator coefficient, $\rho_m$, is assumed to be 0.85\footnote{It is worth noting that the operator coefficient is a subjective parameter and can be determined experimentally for an operator.} and each packet has $b_m = 256$ bits of information. Communication reliability is set to $1-10^{-5}$.
\subsection{Dataset Collection from Teleoperation Prototype}
\begin{figure}
    \centering
    \includegraphics[width=0.44\textwidth]{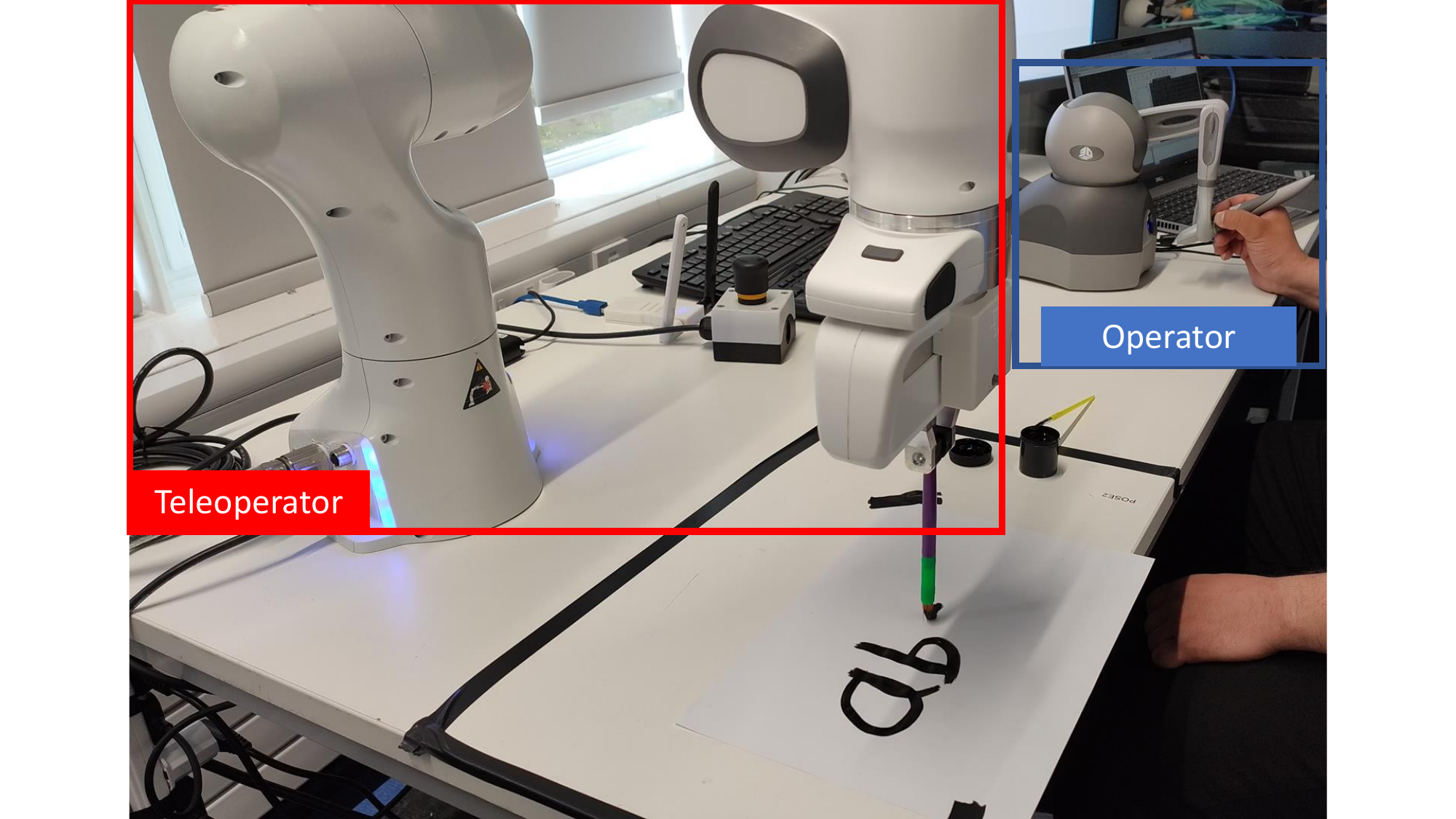}
    \caption{Dataset collection from teleoperation testbed}
    \label{fig:tasks}
\end{figure}

To train task-level predictor, trajectory-level predictor, and DRL agent, we collect real-world trajectory samples from our teleoperation testbed. A human operator controls a robotic arm to write four different letters, i.e., a, b, c, d, as four types of tasks as seen in Fig. \ref{fig:tasks}. Each observation is a time series trajectory given by $\textbf{k}_{t:t^{'}} = \{\textbf{k}_{t}, \textbf{k}_{t+1},...,\textbf{k}_{t^{'}}\}$ where each observation, $\textbf{k}_t = [q_t, v_t, a_t, f_t, tq_t]$, consists of an angular position, $q_t$, angular velocity, $v_t$, angular acceleration, $a_t$, measured external force and torque on the end-effector, $f_t$, and $tq_t$, respectively, in the $t$-th time slot. Each trajectory is collected with timestamps and letter labels. In total, 600 letter trajectory samples are collected which corresponds to $1.98\times 10^5$ data points.
\subsection{Task-level Prediction: User Intention Recognition}
\begin{figure}
    \centering
    \includegraphics[width=0.48\textwidth]{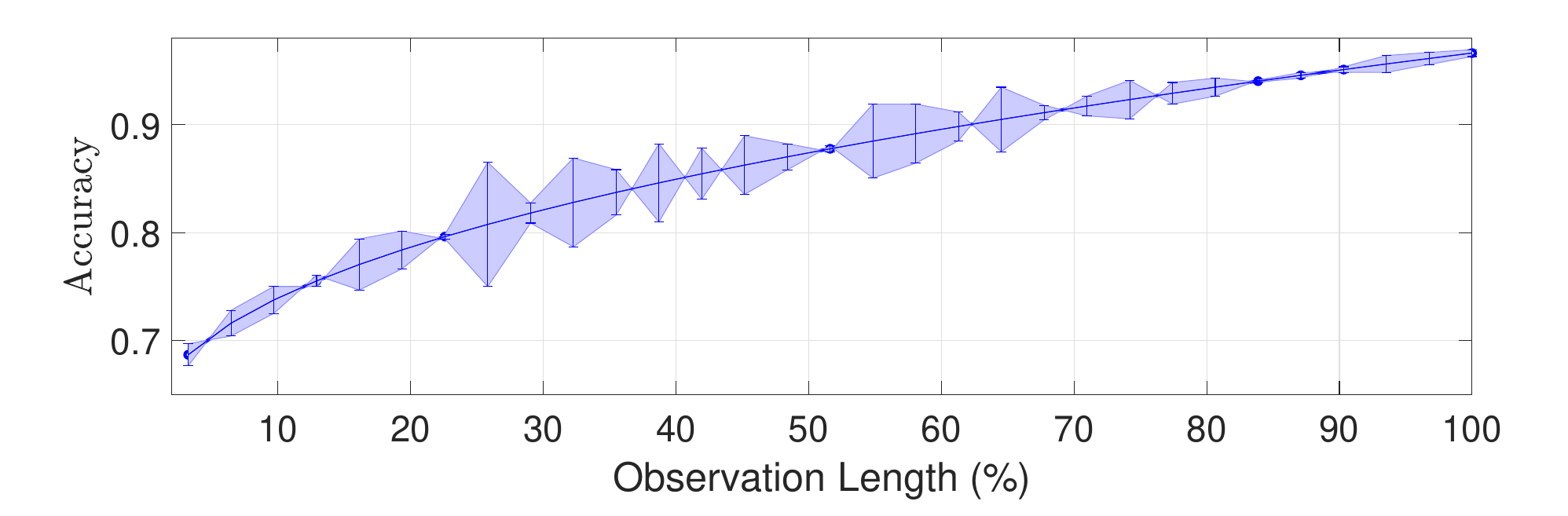}
    \caption{User intention recognition (task-level prediction) accuracy vs observation length (\% of task).}
    \label{fig:acc_vs_obs_len}
\end{figure}

To evaluate the task-level prediction, we conduct experiments with the collected trajectory data. The classification accuracy of task-level prediction is studied with different observation lengths. The accuracy of the task-level prediction increases with observation length, as shown in Fig. \ref{fig:acc_vs_obs_len}. This is reasonable since more information about the task leads to a more accurate task classification. After 60\% of the task, the classification model achieves more than 90\% classification accuracy with smaller deviation as shown in Fig. \ref{fig:acc_vs_obs_len}.
\subsection{Trajectory-level Prediction}
Trajectory-level prediction algorithms are evaluated with different observation lengths between 50\% and 90\%. The observation length of the trajectory-level prediction algorithm depends on mode-switching timings. For example, if the observation length is 50\%, it means that the system is switched to the autonomous mode in the middle of the task and the trajectory-level predictor needs to predict the remaining trajectory to finish the task. Prediction error results for both LSTM and CNN models are provided in Table \ref{tab:pred_accuracy}. According to the results, the accuracy of the trajectory-level prediction increases with increased observation length. This is reasonable since the prediction horizon decreases as the observation length increases. LSTM outperforms CNN, although CNN produces similar accuracy results. This is expected since LSTM performs better on time series (i.e., sequential) data compared to CNN.
\begin{table}
\caption{Performance of Trajectory-level Predictors}
    \centering
    \begin{tabular}{|c|c|c|c|c|}
    \hline
   \textbf{Observation Length (\%)} & \textbf{Errors(\%)}& \textbf{LSTM} & \textbf{CNN}  \\
    \hline
    \multirow{2}{4em}{50\%} & \textbf{Training RRMSE} & $10.66\%$ & $11.04\%$\\
    
    &\textbf{Testing RRMSE} & $10.93\%$ & $11.08\%$\\
    \hline
    \multirow{2}{4em}{60\%} & \textbf{Training RRMSE} & $9.37\%$ & $9.49\%$\\
    &\textbf{Testing RRMSE} & $9.84\%$ & $9.91\%$\\
    \hline
    \multirow{2}{4em}{70\%} & \textbf{Training RRMSE} & $6.61\%$ & $6.94\%$\\
    &\textbf{Testing RRMSE} & $7.48\%$ & $7.68\%$\\
    \hline
    \multirow{2}{4em}{80\%} & \textbf{Training RRMSE} & $4.24\%$ & $5.01\%$\\
    &\textbf{Testing RRMSE} & $4.93\%$ & $5.30\%$\\
    \hline
    \multirow{2}{4em}{90\%} & \textbf{Training RRMSE} & $1.99\%$ & $1.93\%$\\
    &\textbf{Testing RRMSE} & $2.42\%$ & $2.50\%$\\
    \hline
    \end{tabular}
    \label{tab:pred_accuracy}
\end{table}
\subsection{DRL Agent}
\begin{figure}
    \centering
    \includegraphics[width=0.47\textwidth]{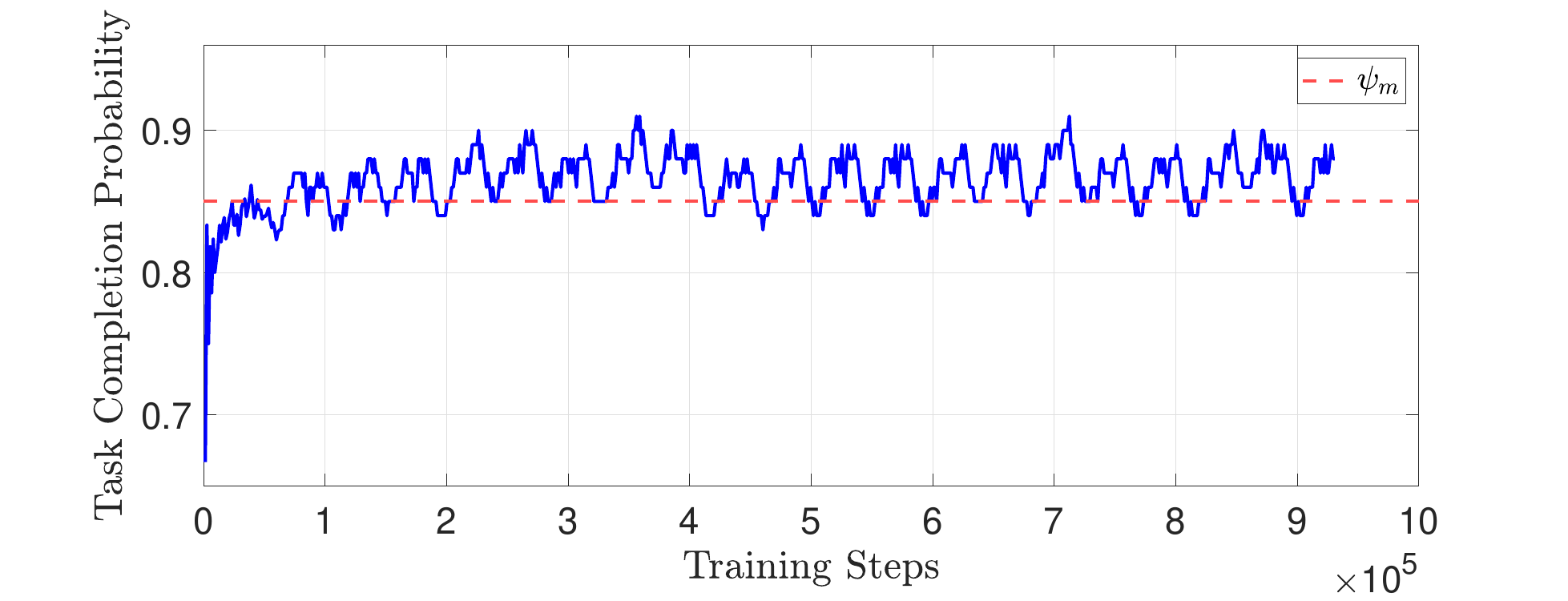}
    \caption{DRL training results for task completion probability, where $\psi_m = 0.85$.}
    \label{fig:rl_success}
\end{figure}

\begin{figure}
    \centering
    \includegraphics[width=0.47\textwidth]{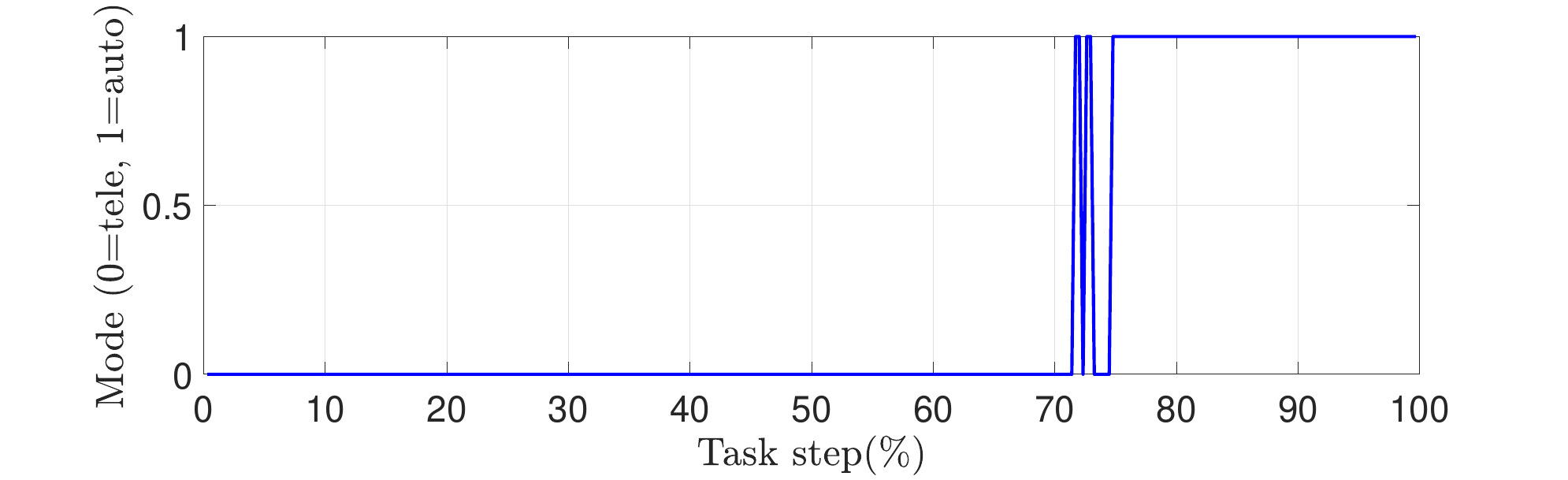}
    \caption{Task example.}
    \label{fig:single_ex}
\end{figure}

In DRL training, we use the proposed task-level and trajectory-level predictors. We provide training results in Fig. \ref{fig:rl_success}. From the figure, task completion probability increases at the beginning and converges to a value after training around 100,000 steps. We further evaluate the agent with a task example, as shown in Fig. \ref{fig:single_ex}. System starts with the teleoperation mode, then the agent switches to the autonomous mode around 70\% of the task. There are multiple mode switches that are triggered by the error detection mechanism. In other words, the agent switches back to the teleoperation mode if there is an error in user intention recognition. Around 75\% of the task, agent switches back to the autonomous mode to finish the task autonomously. As seen from the figure, robot's operating mode switches rapidly and frequently over a period of time before switching to fully autonomous mode. These frequent switches happens in very small amount of time, i.e., $<500$ ms, hence, they are not noticeable by the operator. In other words, frequent and rapid switches over a period of time before switching to fully autonomous mode do not lead to unsmooth robot operation. In brief, results show that the agent learns the dynamics of the system and acts as expected.
\subsection{Overall Results}
\begin{figure}
    \centering
    \includegraphics[width=0.45\textwidth]{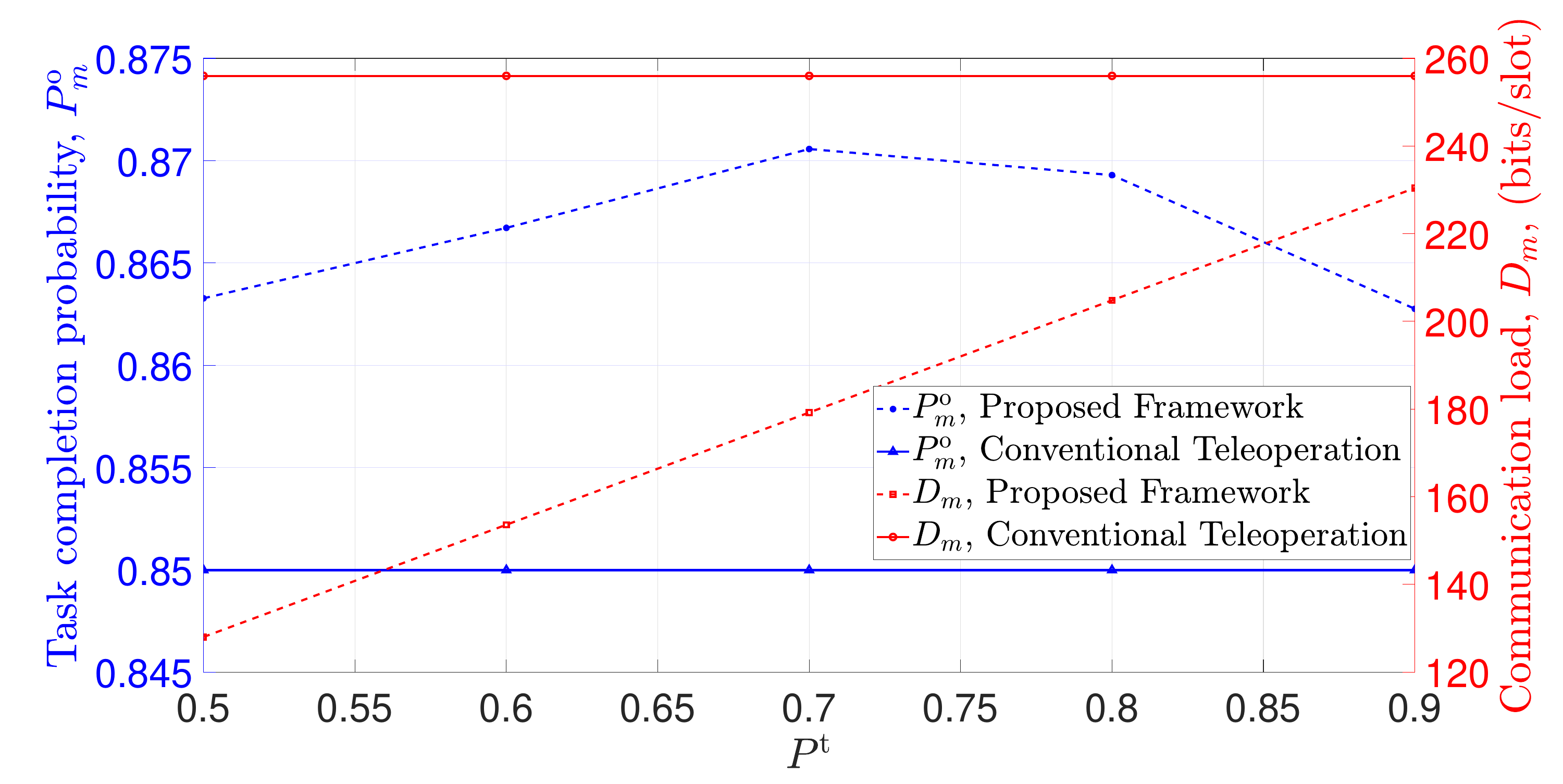}
    \caption{Task completion probability and communication load vs probability that system stays in teleoperation mode, where $\rho_m=0.85$, $b_m=256$ bits/slot, and $\epsilon^{\rm d}_m= \epsilon^{\rm q}_m=10^{-5}$.}
    \label{fig:suc_and_load}
\end{figure}

\begin{figure}
    \centering
    \includegraphics[width=0.46\textwidth]{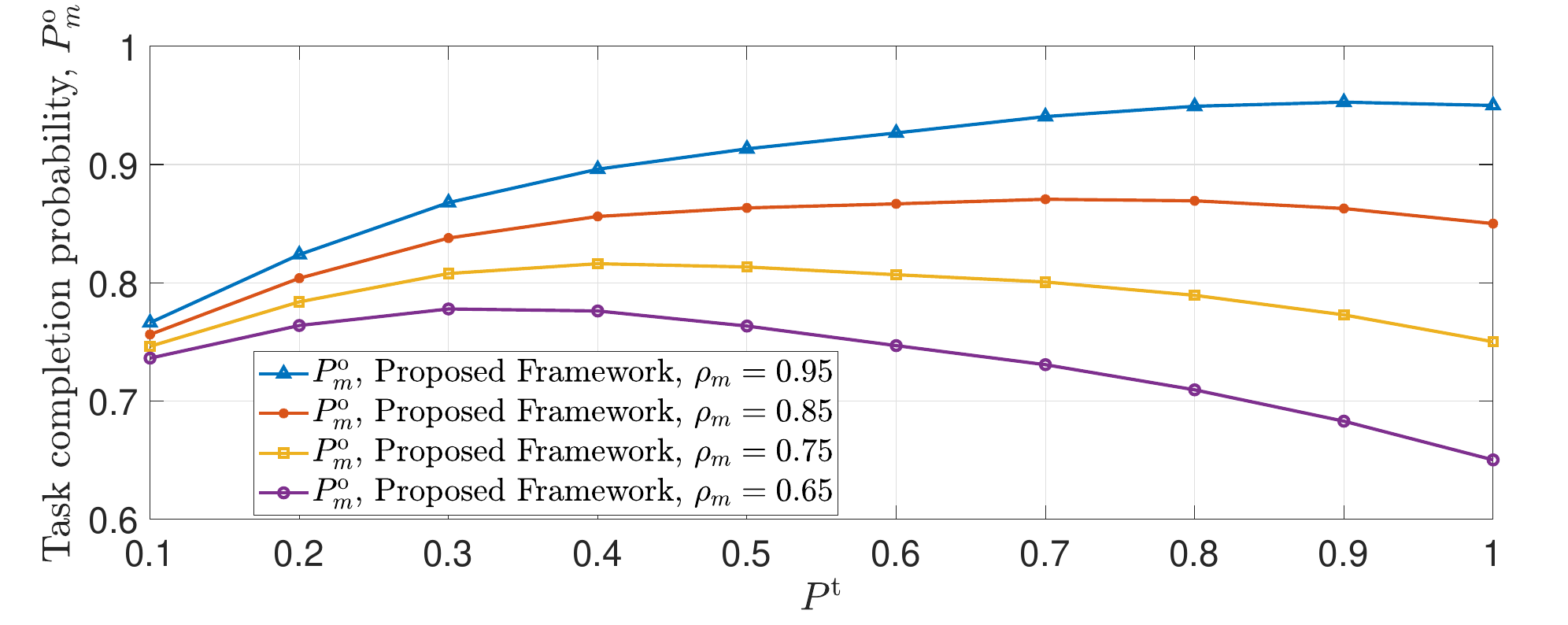}
    \caption{Task completion probability vs probability that system stays in teleoperation mode, for different operator experience coefficient $\rho_m$, where $b_m=256$ bits/slot, and $\epsilon^{\rm d}_m= \epsilon^{\rm q}_m=10^{-5}$.}
    \label{fig:suc_and_op}
\end{figure}

We evaluate the proposed framework by comparing it with conventional teleoperation. In conventional teleoperation, the operator controls the robotic arm to finish the task, and task completion probability depends on the communication reliability and the operator coefficient as formulated in (\ref{eq_succ_tele}). In the proposed framework, intelligent mode-switching is applied and task completion probability depends on the intention recognition algorithm, the error detection system, and the trajectory-level prediction as formulated in (\ref{eq_succ_prop}). According to our results, the proposed framework demonstrates comparable task completion probabilities while significantly reducing communication load, as illustrated in Fig. \ref{fig:suc_and_load}. The figure reveals an interesting trend that task completion probability initially increases and then decreases with the probability that the system is in teleoperation mode, denoted as $P^{\textbf{t}}$. This behavior can be explained through two asymptotic scenarios. First, when $P^{\textbf{t}}$ is too small, the system operates solely in autonomous mode, relying on user intention recognition and trajectory prediction algorithms, which are not entirely error-free. Consequently, task completion probability may suffer due to limited operator input. Conversely, when $P^{\textbf{t}}$ is excessively large, the system remains constantly in teleoperation mode, relying heavily on communication performance and operator experience, both of which are also prone to errors. In such cases, task completion probability may decline due to overreliance on human intervention. To achieve an optimal task completion probability, the system must strike a balance between autonomy and teleoperation, effectively combining the advantages of both approaches. By finding this equilibrium, the proposed framework achieves promising results, maintaining task completion probabilities while significantly reducing communication load. We further illustrate task completion probability for different operator experience coefficient in Fig. \ref{fig:suc_and_op}. We observe similar trend of task completion probability for different user experience coefficient values.

\begin{figure}
    \centering
    \includegraphics[width=0.47\textwidth]{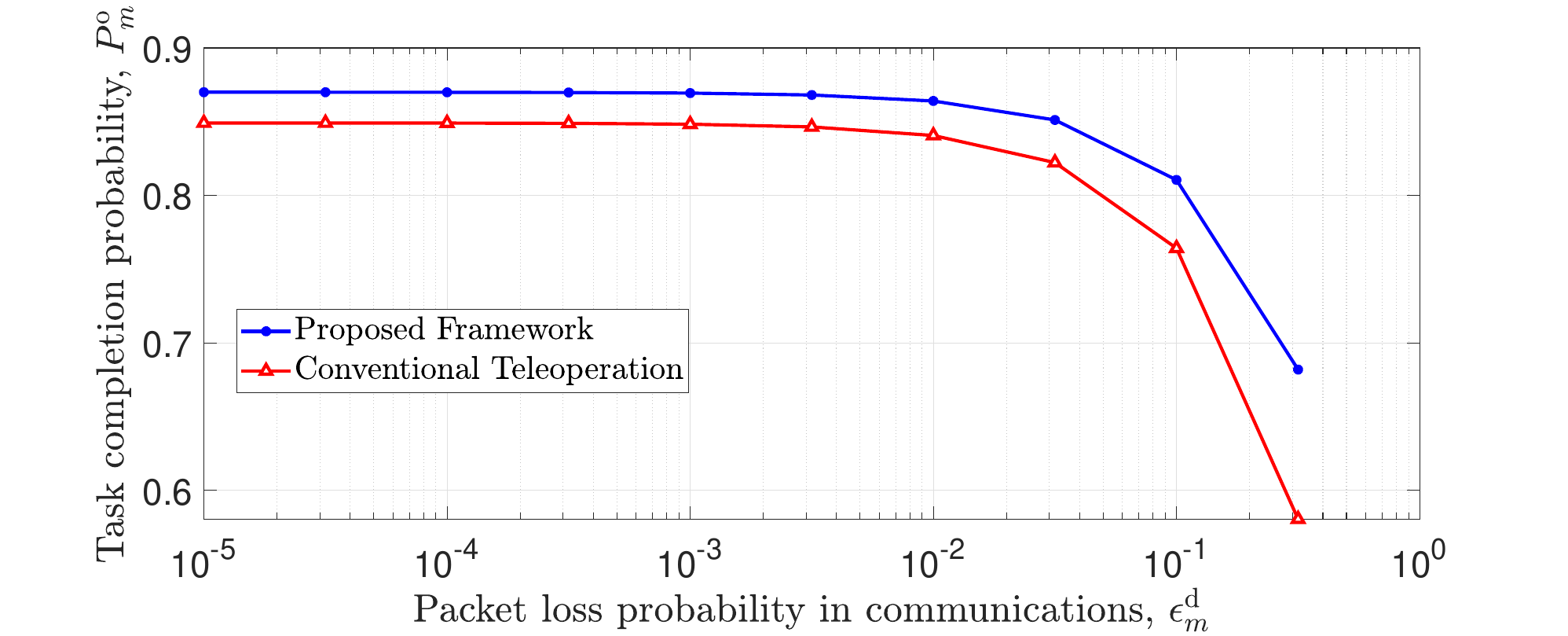}
    \caption{Task completion probability vs packet loss probability in communications, where $b_m=256$ bits/slot, $P^{\rm t} = 0.7$, and $\rho_m=0.85$. }
    \label{fig:succ_vs_comm}
\end{figure}

\begin{figure}
    \centering
    \includegraphics[width=0.47\textwidth]{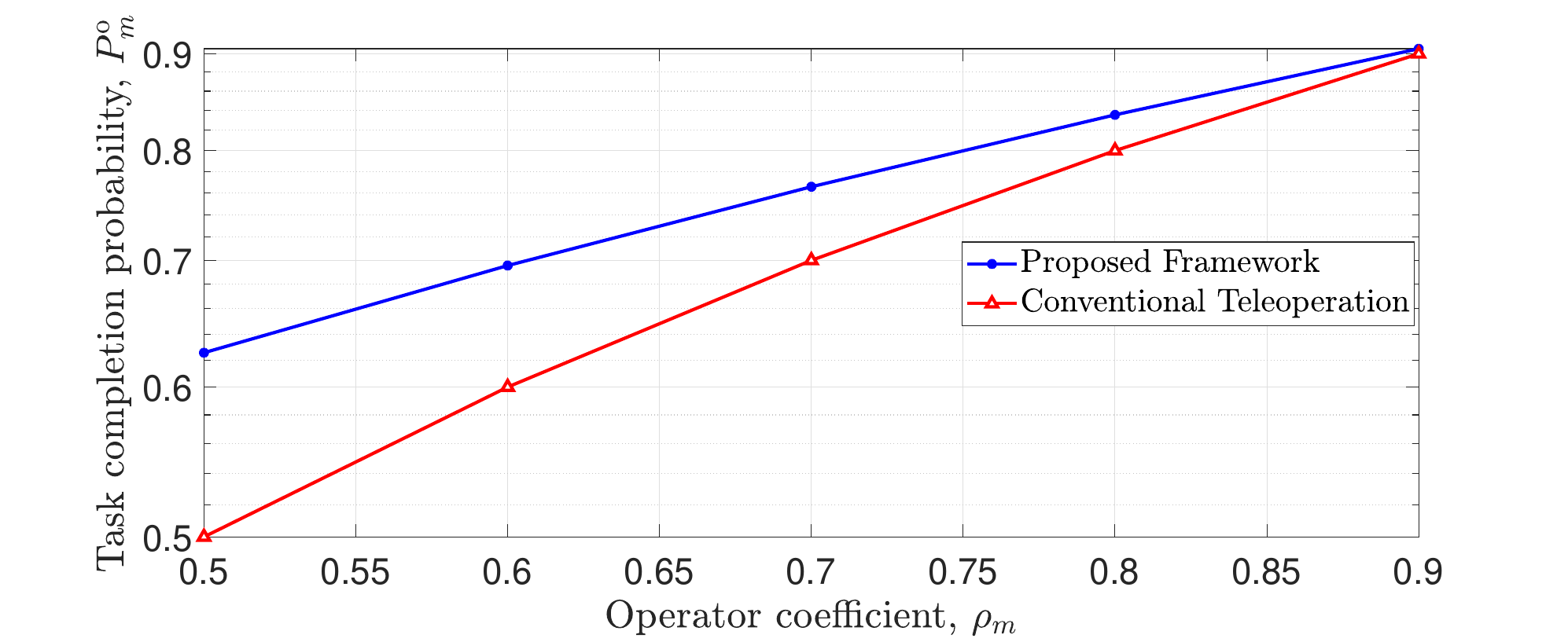}
    \caption{Task completion probability vs operator experience coefficient, where $b_m=256$ bits/slot, $P^{\rm t} = 0.7$, and $\epsilon^{\rm d}_m= \epsilon^{\rm q}_m=10^{-5}$.}
    \label{fig:succ_vs_oper}
\end{figure}

In addition, we provide the comparison between conventional teleoperation and proposed approach in terms of task completion probability for different packet loss probabilities in Fig. \ref{fig:succ_vs_comm}. Both approaches perform similar when the packet loss probability is low, however, the proposed framework outperforms the conventional teleoperation when the packet loss probability is higher, showing that proposed framework is more resilient to poor network conditions. Furthermore, we provide comparison in terms of task completion probability for different operator experience coefficient values in Fig. \ref{fig:succ_vs_oper}. From the figure, both approaches perform similar when the operator experience coefficient is high, however, the proposed framework outperforms the conventional teleoperation when the operator experience coefficient is lower, showing that the proposed framework is more resilient to novice operator. In brief, the proposed framework improves task completion probability for novice operators and under poor communication conditions. 

\section{Conclusions}
We propose an intelligent mode-switching framework to reduce communication load and improve task completion probability. The proposed framework presents an end-to-end joint design of mode-switching and communication systems to enhance the overall performance. A real-world data set is used to train and test the system to validate the feasibility of the proposed framework in realistic scenarios. Our results corroborate that the proposed framework can achieve up to 50\% communication load reduction with improved task completion probability. As a future work, subjective user studies can be conducted to further evaluate the proposed framework in terms of physical and mental demands on the operator. In addition, further comparison with manual switching systems in the literature would be useful to further demonstrate the superiority of the proposed framework.

\bibliographystyle{IEEEtran}
\bibliography{ref}

\end{document}